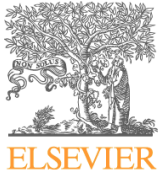

## Pattern Recognition Letters
journal homepage: www.elsevier.com# GPU based Parallel Optimization for Real Time Panoramic Video Stitching

Chengyao Du[a], Jingling Yuan[a, b, ]∗, Jiansheng Dong[a], Lin Li[a, b], Mincheng Chen[a] and Tao Li[c]

[a] *School of Computer Science and Technology, Wuhan University of Technology, Wuhan, 430070, China*
[b] *Hubei Key Laboratory of Transportation Internet of Things, Wuhan University of Technology, Wuhan, 430070, China*
[c] *Department of Electrical and Computer Engineering, University of Florida, Gainesville, 32611, USA***ABSTRACT**

Panoramic video is a sort of video recorded at the same point of view to record the full scene. With the development of video surveillance and the requirement for 3D converged video surveillance in smart cities, CPU and GPU are required to possess strong processing abilities to make panoramic video. The traditional panoramic products depend on post processing, which results in high power consumption, low stability and unsatisfying performance in real time. In order to solve these problems,we propose a real-time panoramic video stitching framework.The framework we propose mainly consists of three algorithms, L-ORB image feature extraction algorithm, feature point matching algorithm based on LSH and GPU parallel video stitching algorithm based on CUDA.The experiment results show that the algorithm mentioned can improve the performance in the stages of feature extraction of images stitching and matching, the running speed of which is 11 times than that of the traditional ORB algorithm and 639 times than that of the traditional SIFT algorithm. Based on analyzing the GPU resources occupancy rate of each resolution image stitching, we further propose a stream parallel strategy to maximize the utilization of GPU resources. Compared with the L-ORB algorithm, the efficiency of this strategy is improved by 1.6-2.5 times, and it can make full use of GPU resources. The performance of the system accomplished in the paper is 29.2 times than that of the former embedded one, while the power dissipation is reduced to 10W.

Keywords—panoramic video; image stitching; embedded GPU; Local ORB; stream parallel optimization2012 Elsevier Ltd. All rights reserved.---

∗ Corresponding author. Tel.: +008613037114558; e-mail: yuanjingling@126.com

## 1. Introduction

Smart city is the product of modern information technology and urban comprehensive management. Its construction is based on the comprehensive perception, automatic analysis and scientific decision-making of various information of the city. Video surveillance is the key and pioneer for smart city construction, play a huge role. It is applied to target tracking[1], person re-identification[2-4], traffic management[5]. With the maturity of high-performance computing[6] and deep learning technology[7-8], and the continuous research and improvement of video processing algorithms through many scholars[9-10], video surveillance technology has developed rapidly.

Panoramic video is the video recording omni-directional scenes at the same point [11,12,13]. Panoramic video has a wide range of applications in video surveillance, robot vision, digital city, live matches, and emerging virtual reality. Panoramic video acquired from panoramic camera shooting is able to bring the immersing interactive roaming, rotating and zoom observing at any angle in the virtual reality [14]. The utilizing of panoramic equipment in military monitoring will improve the battlefield perception of the troops, which enhance the individual combat capability. When it is used in UAV remote sensing and robot vision, the problem of limited view with previous one camera would be solved, which increased the efficiency of detection and recognition

Creating a panoramic image requires a very complex acquisition and stitching process, so the CPU and GPU are required to have a strong processing power [15]. Traditional real-time stitching devices are equipped with processing components such as wide-angle lenses and FPGAs. The devices with low imaging resolution cannot carefully correct images by complex algorithms. In addition, the traditional software-based approach is not real-time [16,17,18], because the information needs to be transferred offline to the stitching software for stitching, after the image is captured by the camera. These present new challenges for scholars. The main work of this paper is as follows:

1) We propose a real-time panoramic video stitching framework.The framework is based on efficient L-ORB image feature extraction algorithm proposed in (2),high accuracy rate feature matching algorithm proposed in (3). In addition, we have effectively utilized the block, thread, and stream parallel strategies of GPU instructions to accelerate this video stitching framework proposed as (4).

2) We propose a L-ORB image feature extraction algorithm, which reduces time cost by optimizing the segmentation of feature detection regions and simplifying the scale invariance and rotation invariance.

3) We propose a feature point matching algorithm based on LSH[19],which use Multi-Probe Locality Sensitive Hashing (Multi-Probe LSH) [20] algorithm to match feature points and utilize the PROgressive Sample Consensus algorithm (PROSAC)[21] to eliminate false matching, for getting the frame image stitching mapping. Then we eliminate the video seams by using the multi band fusion algorithm.

4) The framework accelerates the vedio stitching algorithm by utilizing the block, thread and stream parallel strategy of GPU instruction. Experimental results show that the proposed algorithm can achieve good performance in image stitching in feature extraction, feature matching. So that the performance of the framework proposed in the paper is 29.2 times than that of the former embedded one, while the power dissipation is reduced to 10W.

## 2. Related Work

Generating a panoramic video requires image registration and image fusion for each frame in the video. The key to image registration is feature extraction [22]. Cao et al [23] proposed a speed reflection effective method of extract the edge feature point, the method achieved multi-resolution image fusion stitching. In this method, they build an edge smoothing pyramid and extract the stable features for image registrations. By reusing the multi-scale representation, the registered images are fused, and the cost of mosaic is eliminated. Ethan Rublee proposed ORB algorithm [24] in 2011, adding scale invariant to FAST [25] corner, which solves the problem of rotation invariance and noise sensitivity of BRIEF [26]. The efficiency is 100 times higher than that of SIFT [27,28] algorithm. Jiang et al. [29] proposed a distributed parallel extraction algorithm DP-SIFT, which is based on the characteristics of SIFT feature algorithm. It designs three methods: a high&width limited data block partitioning method, a data distribution method and a feature information adjustment method. Meanwhile they optimized the data block principle and data transmission strategy, greatly reducing the data communication time, and improving the efficiency of the algorithm.Christopher Parker[30] used the ORB algorithm to perform 3D reconstruction in the CUDA parallel computing environment. Chi [31] et al achieved real-time image registration and positioning system based on CUDA, with 20 times faster than CPU, which meet the requirements of real-time processing.

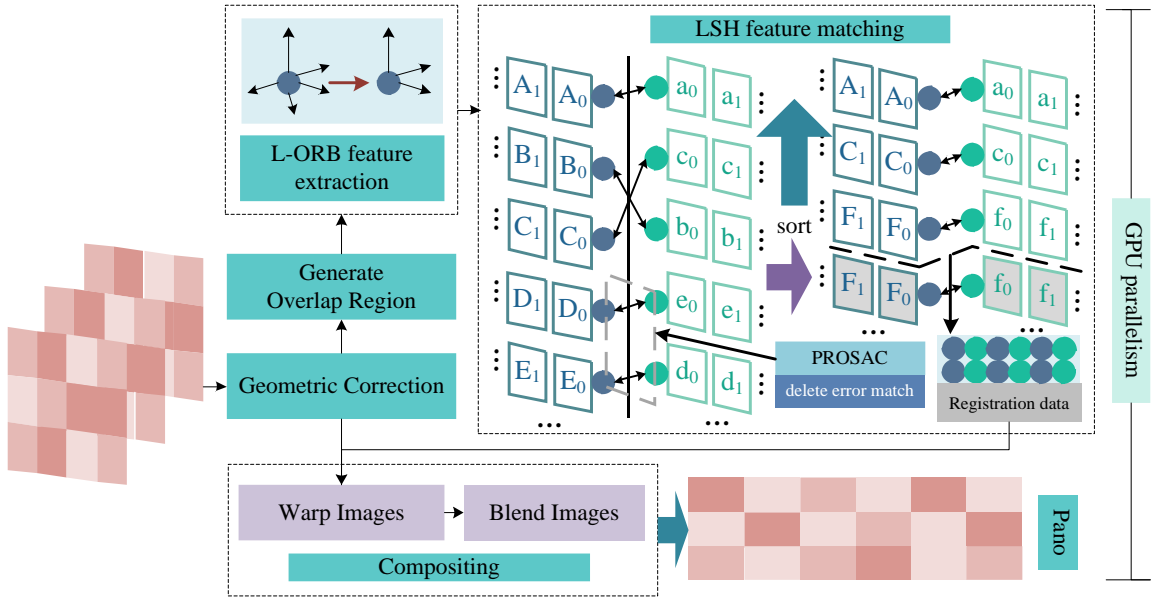

**Fig. 1.** Proposed Framework

## 3. Framework of Vedio Stitching

As shown Figure 1,The framework we propose mainly consists of three algorithms, L-ORB image feature extraction algorithm, feature point matching algorithm based on LSH and GPU parallel video stitching algorithm based on CUDA.By improving the algorithms for image stitching and GPU optimization, the performance of the framework is 29.2 times than that of the former embedded one, while the power dissipation is reduced to 10W.

### 3.1. L-ORB image feature extraction algorithm

In traditional image stitching, the content of the whole picture needs to be done feature point detection and match. Also the matching feature points need to balance the scale invariance and rotation invariance. In the panorama camera group, the relative positions and directions of the camera are fixed, and the space-time complexity of the feature points detection can be reduced by the preprocessing of the image by pre-correcting the parameters.

As shown in the left part of Figure 1,the L-ORB image feature extraction algorithm reduces the detection area by roughly aligning the image with the positional parameters between the cameras. And then the algorithm obtains the distribution range of the feature points by calculating the overlapping field of the image. The L-ORB algorithm combines the FAST feature points with the Harris corner measure method [32]. And it generates the BRIEF feature description factor, which simplifies the scale and rotation invariance relative to the original ORB algorithm, and makes the efficiency greatly improved.

### 3.2. Feature point matching algorithm based on LSH

As shown in the right part of Figure 1,the feature point matching algorithm based on LSH consists of Multi-Probe LSH feature point searching and PROSAC feature point screening, which is more efficient and accurate than the traditional feature point matching algorithm.

In order to cover the most of the neighbor data, original LSH indexes are required to establish multiple Hash Tables, which has high space complexity. Qin Lv, etc. presenting the idea of Multi-Probe LSH algorithm, using a carefully deviated detection sequence to acquire multiple hash buckets approximated to the query data, increasing the possibilities of querying neighbor data.

The traditional RANSAC algorithm has low efficiency, which adopt the random sampling method and neglect the difference between good and bad samples. The PROSAC algorithm sorts the samples by quality to extract samples from higher quality data subsets. After several times hypothesizing and verifying the optimal solution is obtained. Its efficiency is 100 times higher than RANSAC and possess higher robustness.

### 3.3. GPU parallel video stitching algorithm based on CUDA

The proposed L-ORB and LSH algorithms require complex matrix operations on the image, and the CPU's serial processing mode performance cannot meet the real-time requirements. CUDA (Compute Unified Device Architecture) is a parallel computing platform and programming model implemented by NVIDIA's graphics processing unit (GPUs), which has a great acceleration ability of concurrency architecture for a large number of concurrent threads [33,34]. We utilize the GPU's core computing features for implementing the parallel execution of the proposed algorithm. Moreover, we utilize the CUDA architecture to accelerate the concurrent matrix operations, which can double the video splicing speed.

## 4. Three algorithms in our framework

### 4.1. L-ORB image feature extraction algorithm

In traditional image stitching, the content of the whole picture needs to be done feature point detection and match. Also the matching feature points need to balance the scale invariance and rotation invariance. In the panorama camera group, the relative positions and directions of the camera are fixed, and the space-time complexity of the feature points detection can be reduced by the preprocessing of the image by pre-correcting the parameters.

The L-ORB image feature extraction algorithm reduces the detection area by roughly aligning the image with the positional parameters between the cameras. And then the algorithm obtains the distribution range of the feature points by calculating the overlapping field of the image. The L-ORB algorithm combines the FAST feature points with the Harris corner measure method. And it generates the BRIEF feature description factor, which simplifies the scale and rotation invariance relative to the original ORB algorithm, and makes the efficiency greatly improved.

### 4.1.1. FAST Corner Detection Based on Harris Feature

After rough alignment of the image in the overlapped area, the feature points required by image matching do not need scale and rotation invariance. Therefore, we simplify the oFAST and rBRIEF in the ORB algorithm by removing the scale and rotation invariance in order to improve the performance.

FAST feature is a fast feature detection method proposed by Edward Rosten. It does not have the feature of scale invariance, which has an obvious advantage of speed compared with traditional SIFT and SURF methods. The feature points of FAST algorithm detection are defined as follows: If there are enough contiguous pixels in the neighborhood of the pixel P, which are different from the point, the point is considered to be the FAST feature point. As shown in Algorithm 1, steps 1-4 calculate the FAST corner of the input image I. But the FAST feature point does not have the attribute of the corner point. Therefore, steps 5-7 use Harris corner point measurement method to select the N feature points with the largest value of the corner response function from the FAST feature point, where the response function is $R = \det M - \alpha (\text{trace} M)^2$.

**Algorithm 1:** FAST corner detection based on Harris feature

**Input:** image *I*

**Output:** N corner points in the image

1. Calculate the gradient Ix, Iy of the image I (x, y) in directions of X and Y
2. Calculate the product of the gradients of two directions of the image
3. Use the Gauss function to perform the Gauss weighting (take Sigma =1) for Ix2, Iy2, and Ixy so as to generate the elements A, B and C of the matrix M.
4. Calculate the FAST corner point of image I
5. Calculate the Harris response value R of the FAST corner point and delete the corners point less than the threshold t
6. Perform non-maximal suppression and the maximum point in the local neighborhood is the Harris corner point
7. Gets the N corner points with the maximum response value in the Harris corner point

### 4.1.2. BRIEF image feature description factor

Traditional SIFT and SURF features use 128 bit and 64bit floating point data as the feature description factor, which will occupy a lot of storage space and will increase the time of feature matching. BRIEF (Binary Robust Independent Element Feature) uses the gray level relation of the random point pairs in the image neighborhood to establish the image feature description factor, with the characteristics of low time complexity and low spatial complexity.

The establish of BRIEF feature description factor needs to smooth the image first, and then selects nd point pairs through a specific method within a certain range around the feature point. For each point pair (p,q), if the luminance value I of these two points matches I (p)> I (q), the value generated by this point pair is 1; If I (p) <I (q) , then the value in the two-valued string is -1, otherwise it is 0. For all point pairs, we can generate a binary string of nd.

When the rotation angle of the image is greater than 45°, the recognition rate of the BRIEF image feature description factor is almost zero. So it doesn't have the rotation invariance; However, when the rotation angle is within 15°, the recognition rate of the image is greater than 70%, and the lower the rotation angle, the higher the recognition rate; When the rotation angle is very small, the recognition rate is obviously higher than other methods (Figure 2). In the panorama camera group, the rotation angle of aligned overlapping image is only in the range of 0 to 15°. Using the BRIEF algorithm, we can get faster and better results than traditional methods.

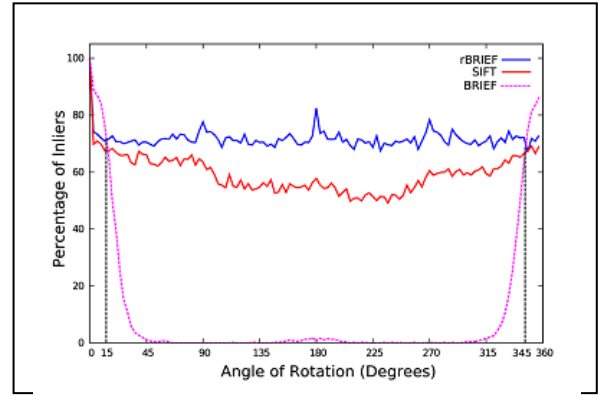

**Fig. 2.** Matching performance under synthetic rotations

As shown in Algorithm 2, the L-ORB algorithm firstly optimizes the feature detection area based on the ORB algorithm by step 1. Then, in step 2-3, the Harris-based FAST corner detection method of Algorithm 1 is used to perform FAST corner detection on the m feature detection regions in step 1 and obtain Harris feature points. Finally, in step 4,the Harris feature points are built into theRIEF image feature description factor.

**Algorithm 2:** L-ORB image feature extraction algorithm

**Input:** n images

**Output:** P groups of feature description factors

1. All n images are divided into m feature detection areas
2. FAST corner detection for m feature detection areas
3. Selecting the P feature points with the highest Harris corner response value from the FAST feature point.
4. Constructing BRIEF image feature description factors with P Harris feature points

### 4.2. Feature point matching algorithm based on LSH

Feature point matching algorithm is to match the feature points of the two images' overlapping area which has same description factors. The probability of matched feature point being adjacent is significant after the images overlapped. Therefore, utilizing Approximate Nearest Neighbor Searching (ANN) algorithm have less both space and time complexity than linear search and k-Nearest Neighbor algorithm (KNN), etc. That could narrow down the searching scope from all feature points to the set of neighbor feature points.

### 4.2.1. Multi-Probe LSH feature point searching

LSH is one of faster methods in ANN. Original LSH utilizes multiple Hash functions Hash mapped vector objects to accomplish dimensionality reduction of data. The final conclusion is obtained through multiple Hash Operation on query vectors and integrating multiple query operations on Hash Table. In order to cover the most of the neighbor data, original LSH indexes are required to establish multiple Hash Tables, which has high space complexity. Qin Lv, etc. presenting the idea of Multi-Probe LSH algorithm, using a carefully deviated detection sequence to acquire multiple hash buckets approximated to the query data, increasing the possibilities of querying neighbor data.

*4.2.2. PROSAC Feature Point Screening*

During the process of images matching, outliers which could cause wrong matches probably occur for any reason. If outliers are introduced in images fusion, it could cause seriously errors. Therefore, outliers must be removed and inliers should be screened to estimate the parameters. In order to keep the following resulting parameter matrixes are closer to the true values, the PROSAC (progressive sample consensus) algorithm is applied in the procedure to eliminate the mismatch points.

The traditional RANSAC algorithm has low efficiency, which adopt the random sampling method and neglect the difference between good and bad samples. The PROSAC algorithm sorts the samples by quality to extract samples from higher quality data subsets. After several times hypothesizing and verifying the optimal solution is obtained. Its efficiency is 100 times higher than RANSAC and possess higher robustness.

Feature point matching algorithm based on LSH consists of Multi-Probe LSH feature point searching and PROSAC feature point screening. Its specific steps are showed in algorithm 3. The algorithm first uses the Multi-Probe LSH algorithm to perform feature point matching on the input feature point set, as shown in step 1. Step 2 then uses the PROSAC algorithm to remove the false match.

**Algorithm 3**: feature point matching algorithm based on LSH

**Input:** A set of feature points for two pictures

**Output**: Screened feature pairs

1. Feature point sets use Multi-Probe LSH for feature point matching

2. Applying PROSAC algorithm to remove error matching

Repeat the following steps until a satisfactory result was found

1) Sorts the matching feature points according to the matching quality from good to poor, and then select the top n higher quality data

2) Randomly choose m data from n data. Calculate the model parameters and the number of inliers.

3) Verify the model parameters

*4.3. GPU parallel video stitching algorithm based on CUDA*

The proposed L-ORB and LSH algorithms require complex matrix operations on the image, and the CPU's serial processing mode performance cannot meet the real-time requirements. CUDA (Compute Unified Device Architecture) is a parallel computing platform and programming model implemented by NVIDIA's graphics processing unit (GPUs), which has a great acceleration ability of concurrency architecture for a large number of concurrent threads. We utilize the GPU's core computing features for implementing the parallel execution of the proposed algorithm. Moreover, we utilize the CUDA architecture to accelerate the concurrent matrix operations, which can double the video splicing speed.

*4.3.1. Blocks, thread parallel*

There are two levels of parallelism between blocks and threads in CUDA functions. Each block is independent of each other, but threads in the same block can exchange data through shared memory. CPU through the priority and time slice to achieve thread scheduling; while, threads on GPU have only two states that are waiting for resources or executed, and it will execute immediately if the resources meet the operating conditions. When the GPU resources are abundant, all threads are executed concurrently, and the acceleration is very close to the theoretical acceleration ratio. When GPU resources less than the total number of threads, some threads will be waiting before the threads executing release resources, which become serialized implementation.

The key to L-ORB algorithm acceleration parallel by CUDA is to divide the serial computing part into multiple subtasks. The image processing procedure such as: FAST feature extraction, non-maximum suppression, the establishment of BRIEF image description factor and image transformation are satisfied the condition which can be divided into multiple sub-tasks. Those sub-tasks have the same calculation process and the data are not related to each other. The parallel design of the panoramic video real-time splicing algorithm on GPU shows as follows:

**Algorithm 4:** panoramic video real-time splicing algorithm

**Input:** multiple cameras simultaneously capture video in different directions

**Output:** N corner points in the image

1. According to the pre-corrected parameters, the video collected for each camera is cut and transformed using the GPU

2. The pixels in the feature detection area are detected in parallel, the pixels matching the FAST feature points are selected and calculated the Harris response values

3. The GPU is used to perform non-maximum suppression of each pixel which meets the FAST feature point

4. The maximum angle P of the Harris corner is obtained, and the BRIEF image feature description factor is established

5. The corners of the two videos are matched and filtered, and the transformation matrix is calculated

6. Use the GPU to transform the video and blend it into a panoramic video

*4.3.2. Stream parallel*

The serial CUDA programming mode for image splicing is divided into three steps: uploading images from main memory to GPU memory, performing corner detection, feature matching, image fusion, downloading the results from GPU memory to main memory (Figure 3a). Application the traditional ORB algorithm in image splicing requires to do a series of things to each picture like: calculate memory usage --> malloc---> Image Stitching --->free memory. These steps will waste a lot of time, result in flow blocking and the processor idle resources. We analyze the image size, the number of picture channels, the number of feature points, the length of describe factor. Moreover, we calculate the required distribution of Grid Size, Block Size, Static Mem and Dynamic Mem. Finally, we implement space pre allocation, when the system starts, which reduce the waste of resources GPU.

CUDA streams divide the program instructions into multiple operation queues, which can achieve the parallel operation between the queues. Since the operation of the different streams is performed asynchronously, we make full use of the GPUs resources by carefully constructing the operation queues which are coordinated with each other. With CUDA stream, the first image can be performed corner detection while the second image is being uploaded, moreover the GPU operation of the multi-image can be detected simultaneously, such as corner detection and feature matching, which greatly saves the time and improves the efficiency (Figure 3b).

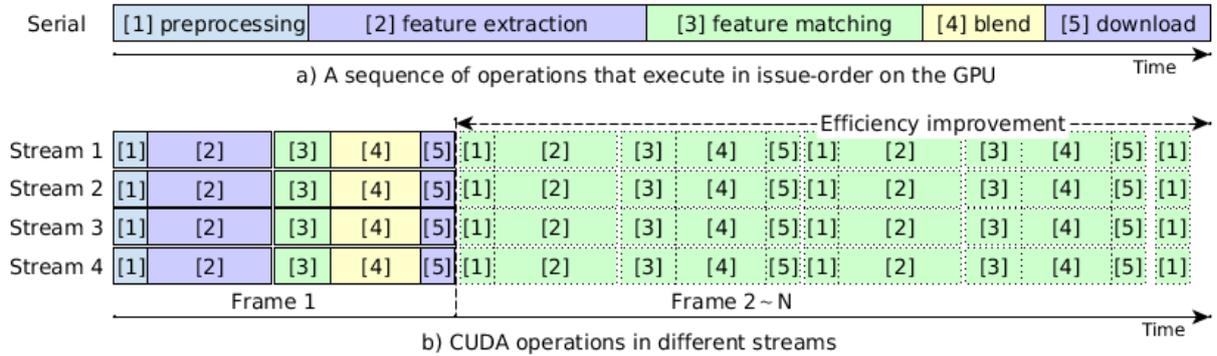

**Fig. 3.** CUDA Stream List

## 5. Experiment analysis

The CUDA and CPU experiments in this paper are running under the Ubuntu 14.04 LTS environment using the Terrans Force X411 computer which is configured for Intel Core i7-6700HQ, RAM 16GB, NVIDIA GeForce GTX 970M; the embedded experiments used the NVIDIA TX2 development board. Software development kits such as NVIDIA CUDA Toolkit 8.0 and OpenCV 3.2.0 are used in the experiment in this paper.

We use SIFT, SURF, ORB, L-ORB, L-ORB + GPU to capture real-time collection of 2304x1728 pixels video and record the stitching time of each stage. Experiments show that the L-ORB + GPU algorithm proposed in this paper has a distinguish performance improvement in the feature extraction and feature matching, etc. whose speed is 11 times faster than classical ORB algorithm and 639 times of classic SIFT algorithm. (Figure 4)

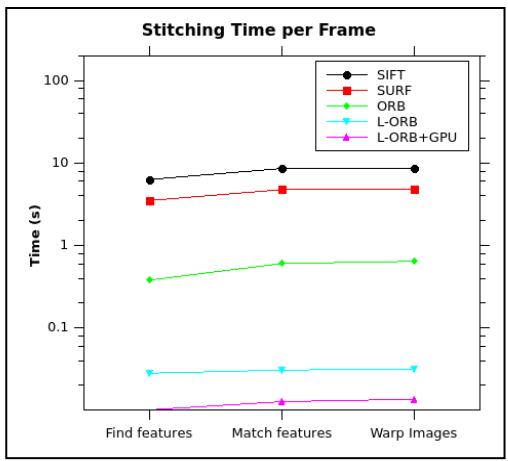

**Fig. 4.** Comparison of Different Algorithms in Three Stages of Image Stitching

In order to verify the effectiveness of CUDA Stream, we utilize NVDIA TX-2 stitching several 1000 frames videos which have different resolution and different number of cameras and calculated the average time cost of stitching every frames. Figure 5a shows the 2 to 7 videos stitching time per frame(s) with using CUDA Stream or not. The result indicates that CUDA Stream has significant acceleration effect on stitching video with a variety of resolution. Figure 5b shows the 2 to 11 videos saving time per frame(s) with using CUDA Stream or not. For the reason that video with different resolutions gathers different contents and extracts different number of corners, including disk I/O and images compression algorithm could influence the stitching time cost, there is no comparison for different resolutions here. The experiment proofs that the performance improves as the video stream increase when the GPU resources are sufficient.

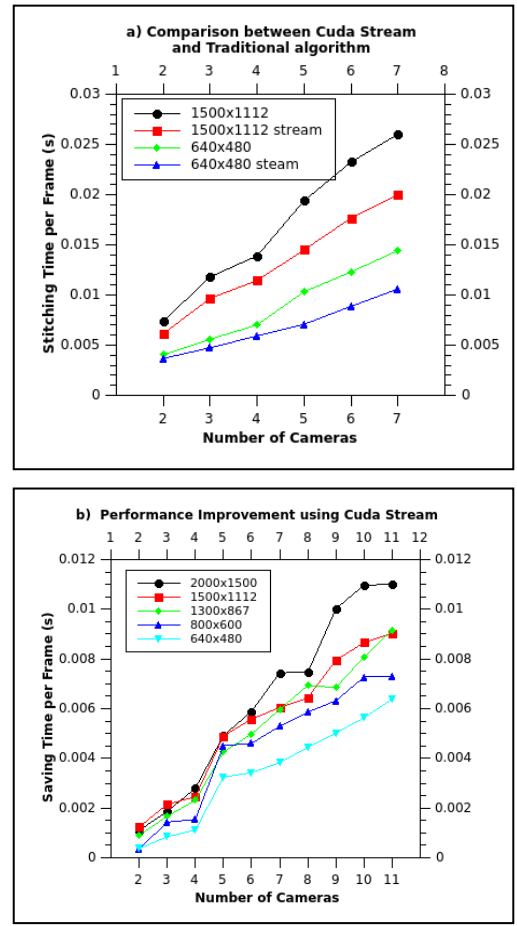

**Fig. 5.** Performance Improvement using CUDA Stream

For verifying the real-time performance of the proposed algorithm, we implement a series of experiments. We use SIFT, SURF, ORB and L-ORB to detect the feature points of video with resolution of 800 * 600, 1020 * 1080 and 2034 * 1728 pixels respectively (Figure 6). In the Intel i7 2.8GHz single thread, L-ORB feature extraction algorithm can reduce the time for the ORB algorithm 1/3, for the SIFT traditional algorithm 1/1000; after CUDA stream parallel, the speed can further enhance 1.6 ~ 2.5 times.

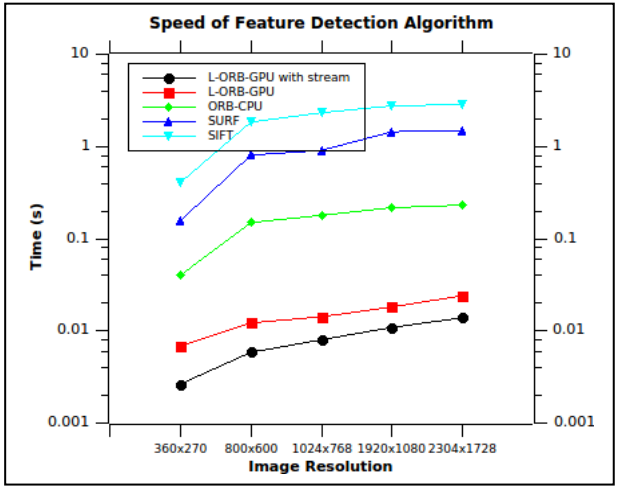

**Fig. 6.** Comparison of Feature Detection Algorithm Efficiency

We use the L-ORB algorithm for image stitching the 2304x1728 pixels data in four GPU embedded like: BeagleBone Black, Raspberry Pi 3B, NVIDIA TX-1 CPU, NVIDIA TX-2. Experimental results show in Figure 7. The algorithm proposed in this paper running in NVIDIA, TX-2 GPU is 29.2 times faster than running in Raspberry Pi 3B, which is the fastest speed of its kind.

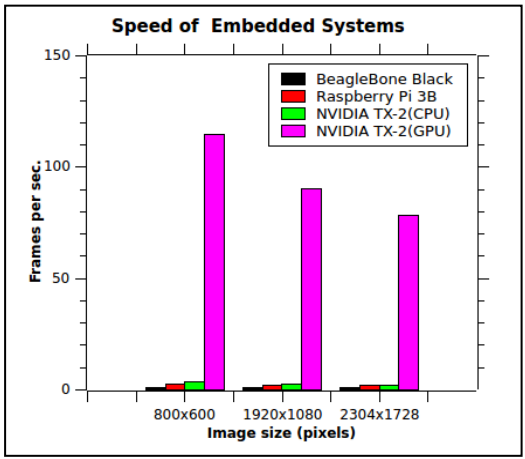

**Fig. 7.** Comparison of image splicing time in embedded devices

## 6. Conclusion and prospect

This paper proposes a L-ORB feature extraction algorithm. The algorithm reduces the feature point detection area, simplifies the support of traditional ORB algorithm on the scale and rotation invariance, which reduces the time complexity of the algorithm. Then we divide the image processing procedure into parallelizable sub-tasks such as: FAST feature extraction, non-maximum suppression, the establishment of BRIEF image description factor and image transformation. Moreover, we utilize the GPU block, thread, flow parallel method for the algorithm to speed up the optimization, which further enhance the efficiency by 1.6-2.5 times of the algorithm. Experimental results show that the method proposed in this paper is over 20 times than traditional ORB algorithm, which can meet the real-time requirement.

Although the experiment can rely on two embedded GPUs to achieve real-time image splicing speed in the experiment, but due to the current computing resources of the chip, the algorithm cannot be fully integrated into one GPU. The future work is to reduce the waste of resources by combining two algorithms in one chip with a more powerful embedded chip.

## Acknowledgments

This research was partially funded by the National Natural Science Foundation of China (Grant No. 61303029).